\documentclass{article}

\usepackage{PRIMEarxiv}
\usepackage{algorithm} 
\usepackage{multirow}
\usepackage{algpseudocode} 
\usepackage[utf8]{inputenc} 
\usepackage[T1]{fontenc}    
\usepackage{hyperref}       
\usepackage{url}            
\usepackage{booktabs}       
\usepackage{amsfonts}       
\usepackage{nicefrac}       
\usepackage{microtype}      
\usepackage{lipsum}
\usepackage{fancyhdr}       
\usepackage{graphicx}       
\graphicspath{{media/}}     

\title{Hammer: Towards Efficient Hot-Cold Data Identification via Online Learning}

\author{
  Kai Lu,  Siqi Zhao, Jiguang Wan\footnotemark[1]\\ Huazhong University of Science and Technology, China \\
  \{\texttt{kailu, m202373801, jgwan\}@hust.edu.cn}\\
}

\begin{document}
\maketitle
\renewcommand{\thefootnote}{\fnsymbol{footnote}}
\footnotetext[1]{Corresponding Author.}

\begin{abstract}
Efficient management of storage resources in big data and cloud computing environments requires accurate identification of data's "cold" and "hot" states. Traditional methods, such as rule-based algorithms and early AI techniques, often struggle with dynamic workloads, leading to low accuracy, poor adaptability, and high operational overhead. To address these issues, we propose a novel solution based on online learning strategies. Our approach dynamically adapts to changing data access patterns, achieving higher accuracy and lower operational costs. Rigorous testing with both synthetic and real-world datasets demonstrates a significant improvement, achieving a 90\% accuracy rate in hot-cold classification. Additionally, the computational and storage overheads are considerably reduced.
\end{abstract}


\section{Introduction}\label{sec_intro}






In the contemporary landscape of big data and cloud computing, the efficient management of storage resources has emerged as a paramount concern. One of the most critical aspects of this challenge is the accurate identification of data's "cold" and "hot" states. Data is classified as "hot" if it is frequently accessed, necessitating fast and readily available storage solutions. Conversely, "cold" data, which is rarely accessed, can be stored more cost-effectively in slower, less expensive storage mediums. Effective hot-cold identification not only optimizes storage costs but also enhances system performance by ensuring that the most relevant data is quickly accessible\cite{bac1, bac2, bac3}.

The importance of hot-cold data identification cannot be overstated. In large-scale data centers, the ability to distinguish between hot and cold data can lead to substantial savings in both storage and energy costs\cite{Int3}. Moreover, it plays a crucial role in improving the overall efficiency of data retrieval processes, thereby enhancing user experience and system responsiveness\cite{Int1}. As data volumes continue to grow exponentially, the need for robust and adaptive hot-cold identification mechanisms becomes increasingly evident\cite{Int2}.

Traditional approaches to cold-hot identification have primarily relied on rule-based algorithms and early artificial intelligence (AI) techniques. Rule-based methods often involve predefined criteria for classifying data as hot or cold, such as the number of accesses within a specific time frame\cite{tra2}. These methods are relatively simple to implement and can be effective in stable environments. However, they suffer from several limitations. Under dynamic workloads, where access patterns change frequently, the accuracy of rule-based methods can significantly degrade\cite{tra1}). Additionally, their lack of adaptability means they cannot easily adjust to new types of data or usage scenarios, leading to suboptimal performance\cite{AI1}.

More advanced traditional strategies incorporate AI to improve upon the limitations of rule-based systems. These AI-based methods typically use machine learning models trained on historical data to predict future access patterns\cite{AI2}. For example, decision trees and neural networks have been employed to classify data based on various features such as access frequency and recency\cite{AI3,AI4,AI5,AI7,AI8}. Despite these advancements, these approaches still face challenges in real-time adaptation and handling large-scale datasets efficiently\cite{AI6}.

\textbf{Problems with traditional strategies.}
One of the primary issues with traditional hot-cold identification strategies is their low accuracy under dynamic loads. As data access patterns evolve, the models used in these strategies may become outdated, leading to misclassification and inefficiencies\cite{AI6, AI7, AI9}. Another significant problem is the poor self-adaptability of these methods. They often require manual intervention to update rules or retrain models, which can be time-consuming and resource-intensive\cite{AI10}. Furthermore, the continuous monitoring and manual configuration adjustments required by these strategies can result in substantial operational overhead\cite{AI6, AI11}.

\textbf{Our solution: an online learning-based approach.}
To address the limitations of traditional methods, we propose a novel solution based on online learning strategies. Our approach leverages the latest developments in online machine learning to dynamically adapt to changing data access patterns without the need for extensive pre-training or manual intervention\cite{online1, online2}. By continuously learning from new data, our method can achieve higher accuracy in identifying cold and hot data, even under highly variable workloads. This dynamic adaptation capability ensures that the model remains up-to-date and effective, regardless of changes in data access patterns\cite{online2}.
Moreover, our solution is designed to be computationally efficient, reducing the operational costs associated with continuous monitoring and manual adjustments. The reduced need for manual tuning and the ability to scale effectively make our method particularly suitable for modern data storage systems.
Finally, we propose a dynamic hot-cold threshold tuning algorithm for hot/cold discrimination under different workloads.
We have rigorously tested our solution using both synthetic and real-world datasets. The results demonstrate a significant improvement (up to 90\%) in accuracy compared to traditional methods, while also showing a reduction in computational overhead. 




Our key results and contributions are summarized as follows:
\begin{itemize}
    \item We analyze and summarize current non-learning (rule-based) and learning-baesd algorithms for hot-cold data identification and show their problems in terms of accuracy (concept drift) and storage overhead (metadata explosion).

    \item We propose a novel hot-cold data identification system, \textbf{\textit{Hammer}}, which improves accuracy by avoiding the concept drift issue through an online learning algorithm and a dynamic threshold tuning strategy. In addition, Hammer reduces the metadata overhead through an online assessment of data hotness based on the Sketch-Min counting method.
    
    \item Thorough experiments demonstrate that Hammer achieves more than 90\% hot-cold identification accuracy across a wide range of application workloads (including AI, HPC, big data, and graph computing) with little additional overhead.
\end{itemize}

The remainder of this paper is organized as follows: Section \ref{sec_mot} reviews related work and issues on hot-cold data identification. Section \ref{sec_design} presents our proposed method in detail, including the overall architecture and algorithmic design. Section \ref{sec_eva} describes our experimental setup and evaluates the performance of our method. Finally, Section \ref{sec_con} concludes the paper.

\section{Motivation}\label{sec_mot}
\subsection{Related Work}

\textbf{Rule-Based Algorithms.}
Traditional hot-cold identification methods often rely on rule-based algorithms, which use predefined criteria to classify data. One of the most well-known rule-based algorithms is the Least Recently Used (LRU) policy. LRU evicts the least recently accessed data items from the cache, assuming that infrequently accessed data is likely to remain cold. While LRU is simple and effective in stable environments, it can struggle with dynamic workloads where access patterns change frequently\cite{LRU}.
Another common rule-based method is the Least Frequently Used (LFU) policy, which tracks the frequency of data access and evicts the least frequently accessed items. LFU is more resilient to temporal variations in access patterns but can be computationally expensive due to the need to maintain and update access counters\cite{LFU1}.

Hybrid approaches combine multiple rule-based policies to improve accuracy and adaptability. For example, the Adaptive Replacement Cache (ARC) combines elements of LRU and LFU to provide better performance under varying workloads\cite{hybrid}. ARC maintains two lists of recently and frequently accessed items and dynamically adjusts the size of these lists based on the observed access patterns. However, hybrid approaches can still be limited by their reliance on fixed rules and may not adapt well to highly dynamic environments.

\textbf{Learned-Based Algorithms.}
With the advent of machine learning, more sophisticated methods have been developed to identify cold and hot data\cite{AI6, AI1}. Decision trees, random forests, and support vector machines (SVMs) have been used to classify data based on features such as access frequency, recency, and file size\cite{AI2}. These models can capture complex relationships in the data and provide more accurate predictions than rule-based methods. However, they often require extensive training on historical data and may not adapt well to new or changing patterns.

Deep learning models, such as neural networks, have shown promise in hot-cold identification due to their ability to learn intricate patterns from large datasets. Convolutional Neural Networks (CNNs) and Recurrent Neural Networks (RNNs) have been applied to predict data access patterns and classify data accordingly\cite{AI13}. These models can handle sequential data and capture temporal dependencies, making them suitable for dynamic environments. However, deep learning models can be computationally intensive and require significant resources for training and inference. This resource intensity can lead to substantial latency, which may not meet the stringent real-time requirements of certain applications.

Reinforcement learning (RL) offers a different approach to hot-cold identification by treating it as a decision-making problem. RL algorithms, such as Q-learning and Deep Q-Networks (DQNs), can learn optimal policies for classifying data based on feedback from the environment\cite{RL1}. These methods can adapt to changing workloads and optimize performance over time. However, RL can be challenging to implement and may require careful tuning of hyperparameters.

\subsection{Limitations of Current Solutions}


\textbf{Accuracy Issues: Concept Drift.}
One of the primary challenges with existing hot-cold identification methods is the concept drift problem\cite{concept_drift, concept_drift2, concept_drift3}. Concept drift occurs when the statistical properties of the target variable, which the model is trying to predict, change over time in unforeseen ways. This is particularly prevalent in dynamic environments where data access patterns can change rapidly and unpredictably.

1) Rule-Based Algorithms: Rule-based methods like LRU and LFU are static and do not adapt to changes in data access patterns. As a result, they can become less accurate over time, especially in environments with frequent and unpredictable changes in workload\cite{rule-based-sys, rule-based-sys2, AI4}.

2) Machine Learning Models: While machine learning models can capture complex patterns in data, they are often trained on historical data and may not generalize well to new or changing patterns. This can lead to a decrease in accuracy as the data distribution evolves \cite{AI2, AI5}.

3) Reinforcement Learning: Reinforcement learning algorithms can adapt to changes in the environment, but they may require significant time to converge to optimal policies, especially in highly dynamic settings\cite{RL2, RL3}.


\textbf{Overhead Issues: Metadata Explosion.}
Another significant challenge is the overhead associated with managing metadata, which can become a bottleneck in large-scale systems.

1) Rule-Based Algorithms: Rule-based methods often require maintaining and updating metadata, such as access counters and timestamps. As the volume of data increases, the overhead of managing this metadata can become substantial, leading to performance degradation\cite{rule-based-sys3}.

2) Machine Learning Models: Machine learning models, particularly deep learning models, can generate a large amount of metadata during the training and inference processes. This includes model parameters, feature vectors, and intermediate results, which can consume significant storage and computational resources\cite{AI3}.

3) Reinforcement Learning: Reinforcement learning algorithms often require storing and processing large amounts of state-action pairs and reward signals, which can lead to metadata explosion. This can be particularly problematic in real-time systems where quick decisions are necessary\cite{RL4}.

\section{Design}\label{sec_design}
\subsection{Hammer Overview}
\begin{figure}[!t]
\centering
\includegraphics[width=0.95\columnwidth]{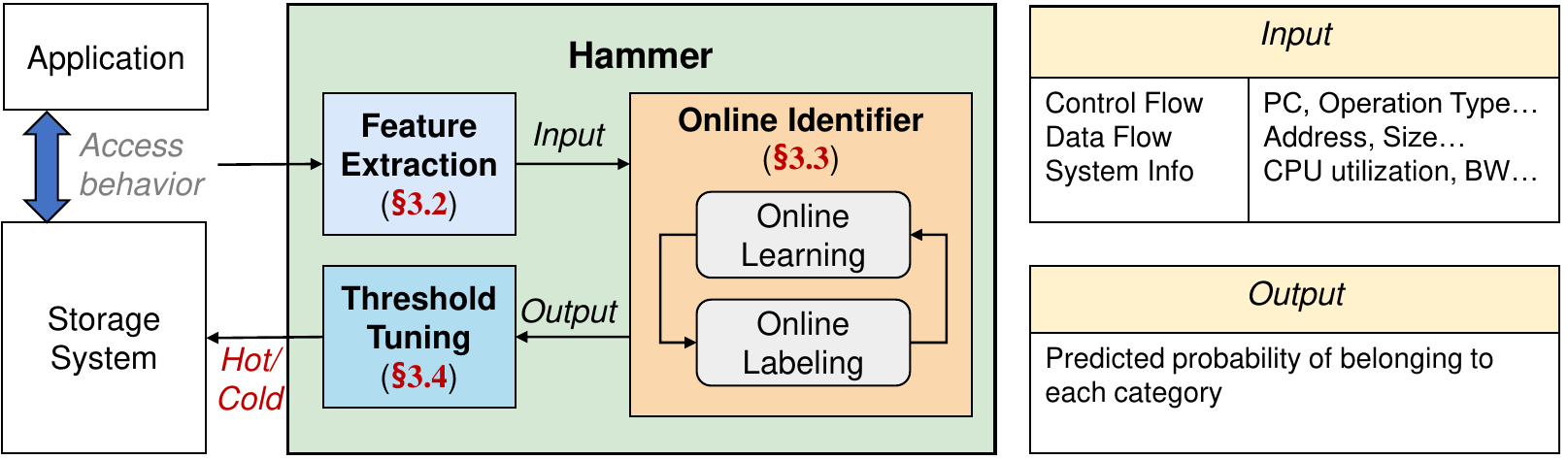}
\caption{Hammer system overview.}
\label{fig-system}
\end{figure}

A data hot/cold identification system has been designed based on the data access behaviour and system state information, as illustrated in Fig. \ref{fig-system}. The system employs online learning algorithms to assess and categorise data as hot or cold in real-time during extensive data processing operations. Additionally, it facilitates multi-path selection for intelligent I/O and optimises data layout for heterogeneous memory or storage systems. The system is structured into three principal modules: feature extraction, data heat prediction and online heat judgement.

\subsection{Feature extraction}
In order to accurately predict the hot and cold states of data, it is essential to extract key features from the storage access behaviour. Such features should encompass not only data flow information, but also control flow information. Furthermore, system information is an essential component. This multi-dimensional feature extraction approach offers a more comprehensive view, enabling the model to comprehend the intricacies of data access behaviour and identify the pivotal factors influencing the hot and cold states of data. The heat feature extraction module constructs feature vectors to characterise data heat based on information such as application behaviour and system state. The information sources are divided into three main categories: data flow, control flow, and system information.

\textbf{Data flow information:} Data flow information serves as the foundation for understanding access patterns. It includes parameters such as the access address, address difference, and operation size. By analyzing these parameters, we can gain insights into access behavior, such as identifying which data are accessed frequently and how they are accessed over time. This information is crucial for predicting data hotness and can help optimize data storage and retrieval strategies.

\textbf{Control flow information:} Control flow information aids in understanding the execution context of data access. It encompasses elements like the program counter (PC), PC difference, and operation type. By examining control flow, we can discern access tendencies, such as whether certain files are accessed sequentially or randomly. This understanding can inform decisions about data caching and prefetching, ultimately improving system performance.

\textbf{System information:} System information provides additional context that influences data access patterns. It includes metrics such as the CPU occupancy of the migration process, bandwidth usage, and the frequency of memory thrashing behavior. These factors can affect how data is accessed and processed, and understanding them allows for better resource allocation and management. For instance, a high CPU occupancy by the migration process, coupled with severe memory thrashing, may indicate data misplacement. In such cases, it suggests that the current distribution of data across the storage hierarchy is suboptimal, necessitating adjustments in data migration strategies to better align data with the access patterns and reduce thrashing. This proactive approach to data management can lead to enhanced system performance and efficiency.

The idea can be applied across various system architectures and environments to optimize data handling and processing. When deployed in communication-intensive scenarios, the algorithm leverages the proximity to the communication endpoint to offload the data collection process. This strategy is particularly effective because it capitalizes on the ease of data acquisition at the communication endpoint, reducing the overhead on the main system and allowing for more efficient use of resources.

\textbf{Data Collection at Communication Endpoints}: By offloading data collection to the near-communication endpoint, the system can take advantage of the low-latency access to data streams. This approach minimizes the need for data transmission back to a central processing unit, thus reducing bandwidth consumption and latency. The algorithm can be tailored to collect specific data access address and size information directly from the endpoint. This information is crucial for understanding the data flow patterns and can be used to optimize data caching and prefetching strategies. The historical memory mechanism introduced by the algorithm ensures that the PC values, which are critical for control flow analysis, are captured at the moment of communication initiation. This mechanism provides a historical context to the data access patterns, enabling the system to make more informed decisions about data management.

\textbf {Application in Heterogeneous Memory or Storage Systems}: In more complex systems, such as heterogeneous memory or storage environments, the algorithm employs the Performance Monitoring Counter (PMC) sampling method for data collection. These systems often involve multiple types of memory or storage media with different performance characteristics, which can make tracking each access behavior computationally expensive and complex. The PMC method allows for efficient monitoring by sampling the access behavior at regular intervals, thus reducing the monitoring overhead. This method is particularly useful in systems where the cost of tracking each access behavior is high, as it provides a balance between the detail of the collected data and the resources required for monitoring.

\begin{figure}[!t]
\centering
\includegraphics[width=0.7\columnwidth]{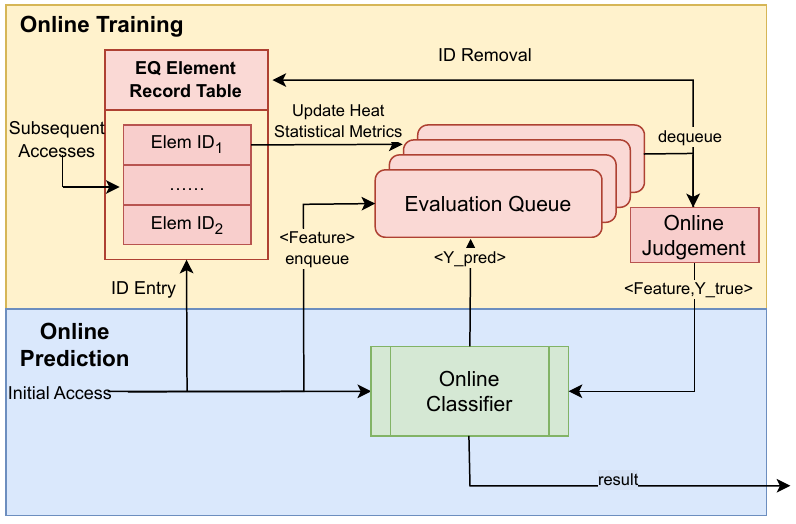}
\caption{Online prediction process.}
\label{fig-learn}
\end{figure}

\subsection{Online Classification and Evaluation}
The design of the online learning classifier is intricately tailored to meet the demands of modern data storage systems. This system is designed to handle a substantial and unceasing flow of requests, necessitating a solution that operates online, processes large volumes of requests continuously, and adapts to the evolving attributes of data over time. The online learning classifier is architected with a online classification and online evaluation process, as depicted in Fig. \ref{fig-learn}.

\textbf{Initial Prediction Stage}: The first stage involves the online classifier generating predicted classification results based on past training as soon as a new data access occurs. This design ensures that the system can provide immediate feedback, which is crucial for real-time data processing systems.

\textbf{Feature Vector Extraction}: Following the initial prediction, the feature vector associated with the new data access is extracted. This vector encapsulates the essential characteristics of the data access, including but not limited to, data flow, control flow, and system information.

\textbf{Evaluation Queue Management}: The extracted feature vector is then added to an evaluation queue, which operates on a first-in-first-out (FIFO) principle. This queue has a finite capacity, ensuring that only the most recent data accesses are retained for evaluation. This design is pivotal for maintaining the system's responsiveness and efficiency.

\textbf{Heat Index Update}: While the feature vector is in the evaluation queue, the system accesses the corresponding part of the memory and updates the heat index. This index is a critical metric that reflects the recency and frequency of data access, thus influencing data storage and retrieval strategies.

\begin{figure}[!t]
\centering
\includegraphics[width=0.75\columnwidth]{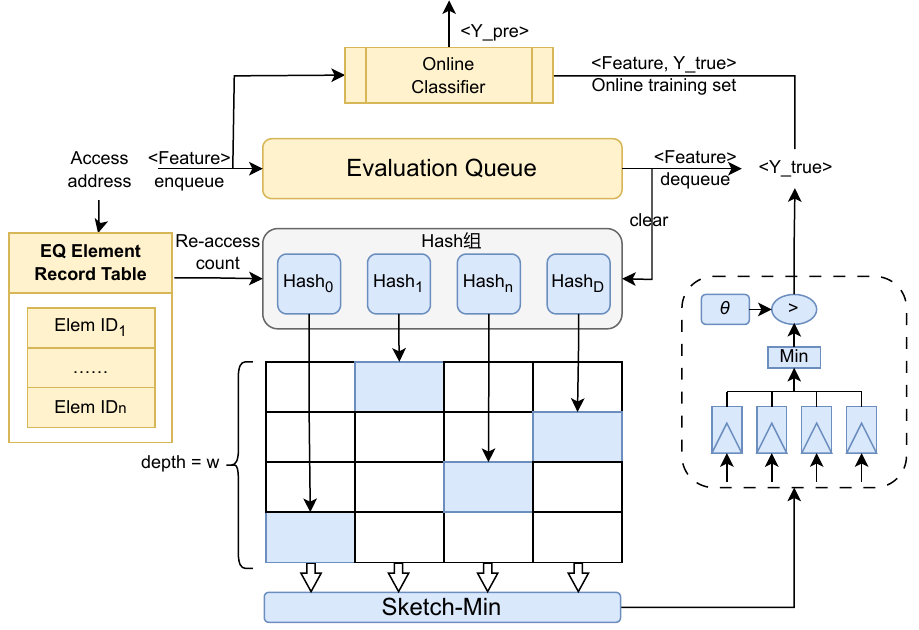}
\caption{Online evaluation strategy based on Sketch-min counting.}
\label{fig-sketch}
\end{figure}

In this study, a Sketch-Min counting-based online method is employed for the evaluation of data heat. The online evaluation algorithm employs a hashing algorithm and a compact data structure to efficiently estimate the number of elements in a high-speed, massive data stream while maintaining a low space overhead. The principal advantage of this technique is its capacity to theoretically guarantee an equilibrium between estimation accuracy and memory usage.

The particular configuration of the online evaluation algorithm is illustrated in Fig. \ref{fig-sketch}. The algorithm achieves the estimation of the number of data accesses by constructing a hash group consisting of D hash functions and allocating W finite hash slots to each hash function. Upon the arrival of a new access record, the count value of each hash function mapped position is increased. Once the access record is removed from the evaluation queue, the actual count value of the element is determined by the minimum of the count values of multiple hash function mapped positions. In the context of large address spaces and massive or even infinite online memory access records, the Sketch-Min counting method effectively controls the total amount of storage, providing an estimate of the number of data accesses with limited storage overhead.

\textbf{Online Training with Real Labels}: Once the queue is full and the feature vector is no longer in the queue, the real label of the data access is obtained through the online evaluation algorithm. This real label is then fed back into the system as training data for the online training of classifiers. This feedback loop is designed to continuously refine the model's accuracy and adapt to changes in data access patterns.

\subsection{Dynamic Threshold Tuning}
In order to respond flexibly to changes in application memory behaviour that are not predictable in advance, we adopt a dynamic hot/cold threshold as a key criterion for determining the hot/cold status of data. The setting of this threshold is based on the percentile value of the hotness of the cells that have recently exited from the evaluation queue and is affected by the usage of the system's slow tier and the memory thrashing behaviour. The setting of the percentile value determines the sensitivity of identifying hot data: the smaller the value, the greater the tendency to identify hot data. The designation of data as hot data may enhance the utilisation of the hot tier, but may also precipitate thrashing issues or an oversupply of the hot tier's resources. Conversely, the designation of data as cold data may help to circumvent thrashing and resource overload, but may also result in the underutilisation of the thermal layer resources.

The precise methodology for adjusting the dynamic threshold is illustrated in Algorithm \ref{alg-dyn}. The initial threshold is established as the ratio of the physical capacity of the hot and cold layers, and the dynamic threshold adjustment mechanism will be triggered at regular intervals as the system operates. In the process of adjusting the threshold, a number of factors are taken into account, including the CPU occupancy of the hot and cold sensing and migration modules, the usage of the slow layer, and the frequency of recent memory thrashing behaviour. The dynamic threshold (i.e. the new percentile) is inversely correlated with the slow tier usage and directly correlated with the memory thrashing behaviour. To quantify this relationship, two empirical coefficients, $\alpha$ and $\beta$, are introduced to characterise the effect of slow layer usage and memory thrashing behaviour on threshold adjustment, respectively.

Furthermore, the dynamic threshold adjustment is constrained by the CPU occupancy of the hot and cold modules. In instances where the hot and cold sensing and threshold adjustment are monitored to be excessive for the CPU occupancy, the system will consider raising the thresholds to reduce the impact on the foreground applications and ensure the stability and responsiveness of the system. This dynamic adjustment strategy enables the hot and cold thresholds to adaptively respond to changes in the system state, thus guiding data storage and migration decisions with greater precision.

\begin{algorithm}[!t]
    \floatname{algorithm}{Algorithm}
    \setcounter{algorithm}{0}
    \caption{Dynamic Threshold} 
    \hspace*{0.02in} {\bf Input:} $P_{min}$: min percentile, $P_{max}$: max percentile, $\theta_{max}$: max CPU utilization  \\
    \hspace*{0.02in} {\bf Output:} p: Hot and cold classification thresholds (percentile) \\
    \hspace*{0.02in} {\bf Default parameter:} $P_{init}$: default percentile, $\alpha$: Slow bandwidth impact factor, $\beta$: ping-pong event impact factor
    \begin{algorithmic}[1]
        \State p := $P_{init}$;
        \State pingpong\_dis := get\_pingpong\_dis()
        \While {dynamic threshold module enabled}
        \State $\epsilon$ := get\_error\_bound()
        \State slow\_band\_rate := get\_band\_rate()
        \State pingpong\_dis\_dif := get\_pingpong\_dis() / pingpong\_dis
        \State cpu\_rate := get\_cpu\_rate()
        \If{cpu\_rate $>$ $\theta_{max}$}
            \State p = min\{$P_{min}$, p/2\}
        \Else
            \State p = p * \(\frac{(1+pingpong\_dis\_dif)^\alpha}{(1+slow\_band\_rate)^\beta}\)
            \State p = min\{p, $P_{max}$\}
            \State p = max\{p, $P_{min}$\}
        \EndIf
        \State Wait for next dynamic threshold period
        \EndWhile 
    \end{algorithmic} 
\label{alg-dyn}
\end{algorithm}

\section{Evaluation}\label{sec_eva}

\subsection{Workloads}
The evaluation of our online learning-based classification system was conducted using a rigorous and comprehensive approach to ensure the accuracy and effectiveness of our algorithm. The dataset utilized in our evaluation was collected in real-world workloads using the Drmemtrace tool, a powerful utility designed for tracing and analyzing memory access patterns in various applications.

\textbf{Data Collection and Preparation.} We instrumented the Drmemtrace tool to attach a statistical module to the target processes, sampling memory access instructions at a rate of 10\%. This approach allowed us to capture a representative subset of the memory access behavior without incurring excessive overhead. The applications selected for data collection spanned four diverse domains: artificial intelligence computations, big data processing, graph computations, and high-performance computing. These domains were chosen for their distinct memory access patterns and the complexity they introduce to data classification algorithms. The details of workloads are shown in Table \ref{tab:Workloads}.

\begin{table}[h]
    \centering
    \caption{Workloads}
    \label{tab:Workloads}
    \begin{tabular}{|c|c|c|c|}
    \hline
    Area & Workload & Dataset & \# Access Cmds \\
    \hline
    \multirow{3}{*}{\centering Artificial Intelligence} & TinyDNN & \multirow{3}{*}{CIFAR-10} & \multirow{3}{*}{73,118,235} \\
    \cline{2-2}
                           & ResNet &                       &                       \\
    \cline{2-2}
                           & YOLO &                       &                       \\
    \hline
    Big Data Processing & MapReduce-WordCount & Wiki & 109,541,354 \\
    \hline
    Graphical Calculation & PageRank & 1m nodes & 64,576,324 \\
    \hline
    HPC & Distance-Matrix & 30 dims * 100k vecs & 26,696,931 \\
    \hline
    HPC & Range-Query & 500k regions * 20k queries & 54,134,885 \\
    \hline
    \end{tabular}
    \label{tab:my_label}
\end{table}

\textbf{Simulation and Algorithm Assessment.} To simulate a system running multiple workloads continuously, we concatenated the memory access behaviors collected from all categories into a continuous record. This method was employed to validate the algorithm's ability to adapt to concept drift, a phenomenon where the statistical properties of the data change over time. Using a custom simulator, we replayed the collected memory access records and conducted online predictions to evaluate the accuracy of our hot and cold data classification algorithm. We employed a rigorous evaluation protocol that included precision and F1 Score as primary metrics. For establishing a baseline, we used the Least Recently Used (LRU) 2Q algorithm, a well-established method for differentiating between hot and cold data based on the presence of access addresses in the LRU hot queue. This baseline served as a reference point to assess the performance of our online learning-based classifiers.

\subsection{Evaluation Results}

We implemented and tested several classifiers based on classic online learning algorithms. Notably, we included a probabilistic Bayes classifier, a decision tree-based Hoeffding adaptive tree, and an Adaptive random forest (ARF) classifier. Among these, the ARF classifier demonstrated exceptional performance in terms of accuracy and F1 Score. The result is shown in Table \ref{tab:acc}.

\begin{table}[h]
    \centering
    \caption{Accuracy Comparison}
    \begin{tabular}{|c|c|c|}
        \hline
        \textbf{Algorithm} & \textbf{Accuracy} & \textbf{F1 Score} \\
        \hline
        Base: LRU2Q & 71.09\% & 54.70\% \\
        \hline
        Naïve Bayes & 71.21\% & 24.73\% \\
        \hline
        Hoeffding adaptive tree & 83.04\% & 68.26\% \\
        \hline
        Adaptive random forest & 90.33\% & 86.28\% \\
        \hline
    \end{tabular}
\label{tab:acc}
\end{table}

The ARF classifier's superior performance can be attributed to its ability to adapt to the evolving data characteristics in real-time. Unlike traditional batch learning methods, which require periodic retraining to adjust to concept drift, the ARF classifier continuously updates its model with each new instance, ensuring that it remains highly responsive to changes in the data.

To further validate our findings, we performed a series of cross-validation tests, dividing the dataset into multiple folds and training the classifiers on different subsets. This approach allowed us to assess the generalizability of our classifiers and their robustness to variations in the data.

We employed statistical methods to determine the significance of the performance differences observed between the classifiers. Using paired t-tests, we compared the mean accuracy and F1 Score of the ARF classifier against the baseline LRU 2Q algorithm. The results indicated that the performance improvements were statistically significant, with p-values well below the conventional threshold of 0.05.

To underscore the efficacy of online learning in the context of data storage systems, we conducted a thorough comparative analysis against traditional batch learning methodologies. The traditional batch learning approach was tested by first dividing the collected dataset using an 80-20 split method, where 80\% of the data was used for training and the remaining 20\% for testing.

The comparative results between the online and batch learning models are compelling. The online learning model demonstrated superior performance in terms of Accuracy and F1 Score, which are standard benchmarks for classification models . The batch learning model, while achieving high initial Accuracy and F1 Score, showed a decline in performance as the data distribution evolved over time. This decline was only arrested when the model was retrained with updated data, which is a time-consuming process .The result is shown in xxx.

The results of our testing revealed a significant difference in performance between the online and batch learning models. The online ARF classifier demonstrated a higher Accuracy and F1 Score, which was consistently maintained as new data was introduced. This was particularly evident in scenarios where the data exhibited concept drift, where the online model was able to adapt and maintain its performance, whereas the batch model showed a decline in performance until it was retrained.

\section{Conclusion}\label{sec_con}

Our approach, leveraging online learning techniques, is designed to transcend the constraints of conventional methodologies by swiftly adjusting to evolving data access patterns. This dynamic adaptation allows our solution to maintain superior accuracy while also minimizing operational expenditures. Extensive evaluations utilizing a mix of synthetic and real-world datasets have confirmed a noteworthy enhancement, with an achieved accuracy rate of 90\%. Our method has shown to be more cost-effective, with lower operational expenditures, positioning it as an attractive option for contemporary data storage systems.

\section{Acknowledgement}
This work was sponsored by the National Key Research and Development Program of China under Grant No.2023YFB4502701, the National Natural Science Foundation of China under Grant No.62072196, the Key Research and Development Program of Guangdong Province under Grant No.2021B0101400003, and the Creative Research Group Project of NSFC No.61821003.

\appendix

\bibliographystyle{unsrt}  
\bibliography{hammer}  

\end{document}